# Script-Agnostic Language Identification

**Milind Agarwal, Joshua Otten, and Antonios Anastasopoulos**
George Mason University
{magarwa,jotten4,antonis}@gmu.edu

## Abstract

Language identification is used as the first step in many data collection and crawling efforts because it allows us to sort online text into language-specific buckets. However, many modern languages, such as Konkani, Kashmiri, Punjabi etc., are synchronically written in several scripts. Moreover, languages with different writing systems do not share significant lexical, semantic, and syntactic properties in neural representation spaces, which is a disadvantage for closely related languages and low-resource languages, especially those from the Indian Subcontinent. To counter this, we propose learning *script-agnostic* representations using several different experimental strategies (upscaling, flattening, and script mixing) focusing on four major Dravidian languages (Tamil, Telugu, Kannada, and Malayalam). We find that word-level script randomization and exposure to a language written in multiple scripts is extremely valuable for downstream script-agnostic language identification, while also maintaining competitive performance on naturally occurring text.[1]

## 1 Introduction

In many natural language processing (NLP) tasks or data creation efforts, we often need to first identify the source language of a particular text. For instance, automated translation, part-of-speech (POS) tagging, and web scraping for data collection must typically identify the text's language before performing the given task. The languages involved might occur in non-standard scripts, but as we show in this paper, modern systems are heavily script-dependent in language identification (langID). The result is that most current methods are unable to account for languages written in non-standard scripts. Moreover, script diversity is especially common in low-resource languages. Many bilingual communities choose to write their minority language in the region's dominant system (such as those in Pakistan, Iran, China), instead of their language's traditional writing system (Ahmadi et al., 2023a). It is also common for larger standardized languages to be romanized on the internet and in social media. Finally, some languages simply do not possess one standard script, and are written in multiple writing systems. For instance, the Western-Indian Konkani language is actively written in up to 5 scripts: Devanagari, Romi, Kannada, Malayalam, and Perso-Arabic (Lehal and Saini, 2014; Rajan, 2014). However, most Konkani systems only support Devanagari and Romi scripts, and would not recognize the language if written in the other three. This illustrates the need to have script-agnosticism so we can collect high-quality data for low-resource languages, and support their script-diverse nature in NLP applications.

Script-agnostic langID is expected to be most useful for closely related languages that currently do *not* use the same script and where languages often have unique scripts - a scenario most commonly occurring in the Indian Subcontinent. In this paper, we conduct a case study on script-agnosticism for language identification by focusing on the four major Dravidian languages: Tamil, Telugu, Kannada, and Malayalam. We explore three different methods of training script-agnostic embeddings, evaluate on the langID task across domains, and offer insights for future work. Broadly, we attempt to answer the following research questions:

1. What impact does training on transliterated corpora have on downstream langID?
2. How does projecting to one script or upscaling to multiple scripts impact performance?
3. What impact does intra-sentence script mixing have on language identification?

[1] Data, models and code available here: https://github.com/Joshua-Otten/Script-Agnostic-Lang-ID

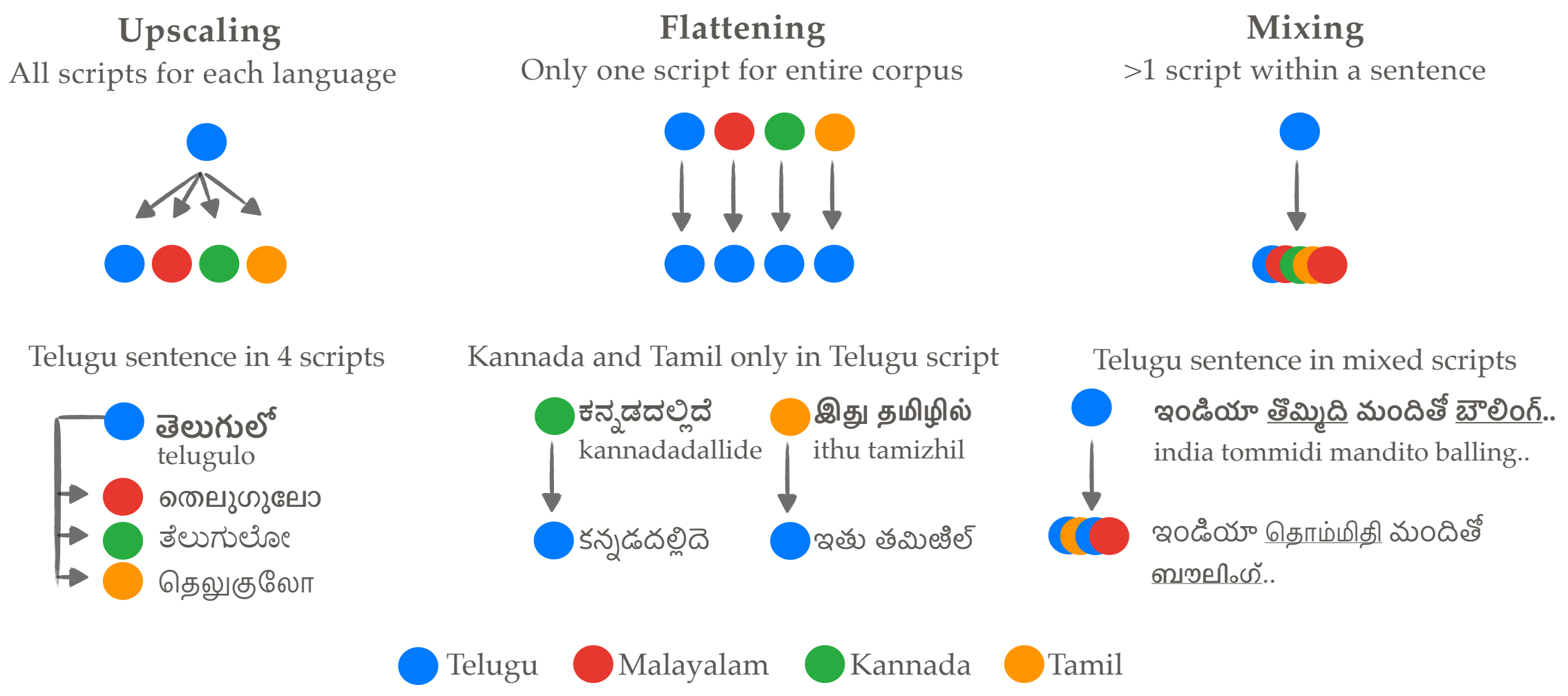

Figure 1: In upscaling, we transliterate each sentence into other scripts to expose the model to data in that language in all 4 writing systems. For flattening, we aim to reduce this potential vocabulary overload and project all scripts into 1 script per experiment. The goal is to identify *if* any of the 4 scripts is a suitable target script for all languages. Each writing system has a unique number of total letters even though there is large overlap (Table 1), and we think that this *may* result in one or the other script to be a suitable script for projection. For the final mixing setup, we transliterate at the word-level instead (at different noise levels) and allow multiple scripts per sentence.

## 2 Methods

**Script Flattening** Under this setup, we want to explore whether the embedding space will benefit from seeing all the languages in only *one* common script (Figure 1). The idea behind flattening the script space from four to one is that with only one script, the embedding space (and consequently the classification system) can focus on finding discriminative features between the languages. It is worth noting that training word representations in a single script may perform poorly in real-world settings and may not be a practical choice since text will naturally appear in different scripts and will require transliteration as preprocessing. However, this experiment is useful to quantify the role that script plays in language identification, compared to the non-visual distinguishing features of the languages.

**Script Upscaling** This method takes a given training example written in one script and "upscales" it into all 4 scripts (Figure 1). Our intuition is that seeing every example in each script will prevent a model from giving weight to any one writing system in its decision-making, forcing it to rely on inherent features of the language. In other words, we teach the model that a sentence of a given language could be written in any script, so that it learns not to discriminate on the basis of writing system. This contrasts with the approach taken in Brown (2012), where each language-script pair is given a unique *language model* and their scores are used to make the final classification decision. For our setup, we first created four training files for each language, where a file would include all of the language's training examples four times–one for each script. Then we concatenated all of these files into *one* training set. Therefore, our model assumes that a sentence may appear in any of the four writing systems with the same likelihood.

**Noisy Multi-Script Setup** In the final setup, we create synthetic sentences following Algorithm 1 (Appendix C) and Figure 1 for both FLORES200 data splits. Under this approach, for each noise level *n*, language *lang*, and sentence *sent*, we choose a base script and then randomly pick *n*% words to transform to new non-base scripts. We train separate text classification `fastText` models on each of these noisy datasets and evaluate them on test sets with clean, noisy, and merged datasets. This is to evaluate out-of-distribution generalization and robustness, and the potential usefulness of including noise during the training process. We perform this experiment with permutations of 25%, 50%, 75%, and 100% script-noise levels in the training data as done in Ahmadi et al. (2023b). Finally, we train an "All-Noise" model on merged data from all these script-noise levels.

| Language | 639-3 | Family | Script | Script Code | Vowels | Consonants |
|---|---|---|---|---|---|---|
| Tamil | `tam` | Southern | தமிழ் | Taml | 12 | 18 |
| Kannada | `kan` | Southern | ಕನ್ನಡ | Knda | 16 | 35 |
| Telugu | `tel` | South-Central | తెలుగు | Telu | 16 | 36 |
| Malayalam | `mal` | Southern | മലയാളം | Mlym | 15 | 42 |

Table 1: A summary of the characteristics of the four Dravidian languages we study in our experiments. All four languages use abugidas (alphasyllabaries) for writing and are written from left to right with diacritics. Note that Tamil has the fewest overall graphemes whereas Malayalam has the most. The last two columns indicate *common* vowels and consonants in the language, but each script comes with extended grapheme sets to accomodate other Indian-language phonemes.

| IPA | ISO | TEL | KAN | MAL | TAM |
|---|---|---|---|---|---|
| /ka/ | ka | క | ಕ | ക | க |
| /k$^{h}$a/ | kha | ఖ | ಖ | ഖ | க$_2$ |
| /ga/ | ga | గ | ಗ | ഗ | க$_3$ |
| /g$^{h}$a/ | gha | ఘ | ಘ | ഘ | க$_4$ |

Table 2: Tamil has only one letter to represent the above-mentioned 4 sounds common in the other 3 Dravidian languages. So, the transliterator introduces subscripts to differentiate the four sounds in the source script. There are 5 such character series but we only show the *velar* phonemes' series.

| Model | Acc | N |
|---|---|---|
| CLD3 (Salcianu et al., 2020) | 0.98 | 101 |
| langid.py (Lui and Baldwin, 2012) | 1.0 | 97 |
| Franc[3] | 1.0 | 419 |
| fastText (Joulin et al., 2017) | 1.0 | 176 |
| HeLI-OTS (Jauhiainen et al., 2022) | 0.99 | 200 |

Table 3: This table shows different popular language identification systems, their accuracies on `FLORES-200` and the number of supported languages (N). We chose `fastText` as our root model since it achieves a high accuracy, supports many languages, and can be easily trained from scratch.

## 3 Experiments

**Dataset and Languages** We use the FLORES200 dataset (Costa-jussà et al., 2024; Goyal et al., 2022; Guzmán et al., 2019) for training and in-domain testing in all our experiments. In order to ensure that our models would work well on test data that was not simply from FLORES200, we also tested on three out-of-domain sets: GlotStoryBooks (Kargaran et al., 2023), UDHR (Kargaran et al., 2023), and MCS-350 (Agarwal et al., 2023). We do not transliterate these datasets since the goal is to measure potential performance drops on naturally occurring text compared to traditional models. We also use a subset of monolingual data from IndicCorp (Kakwani et al., 2020) for an experiment involving non-parallel training in §4.2. For this paper, we explore script-agnosticism for 4 major languages (Table 1) that fall within the same language family and use four distinct writing systems. Details about each of the datasets are available in Appendix A and language profiles in Appendix B.

**Transliteration** We use the Aksharamukha[2] python package to transliterate between our four writing systems. Since the library is primarily designed for Indic writing systems, it provides an extremely low-loss 1:1 transliteration, which is suitable for our purposes. This 1:1 mapping is possible because Indic writing systems descend from a shared ancestor - Brahmi script, and they all have unique and mappable graphemes for different phonemes. The only exception (across all Indian writing systems) is Tamil script, which also descends from Brahmi, but in its modern form, it uses one grapheme to represent aspirated and unaspirated or voiced and unvoiced versions of a sound. Aksharamukha adds subscripts (see Appendix Table 2) to differentiate these sounds, but we remove them during preprocessing as they are only found in Tamil writing in literary and classical settings.

**Training Model Choice** Table 3 shows the performance of commonly used off-the-shelf langID models on the FLORES200 dev set. Out of the three highest performing models (with $F_1$ score of 1.0) in Table 3, only `fastText` (Bojanowski et al., 2017) and langid.py can be trained with custom files. `fastText`, trained on data from Wikipedia, Tatoeba and SETimes, supports a wider number of languages in its base model, is known to work well with unknown words, and is very easy to train. Therefore, `fastText` will serve as the training model for our experiments. Macro $F_1$ score is computed across all 4 languages to identify a system with the best overall coverage and accuracy. Moreoever, `fastText` provides an efficient way to glean subword information and is known to better handle out of vocabulary words. Like all other language identification systems, it does not come with script-agnosticism support. All experiments were run on CPU due to fastText's optimizations.

[2] https://pypi.org/project/aksharamukha/

**Training and Evaluation** We obtain our results based on the original versions and transliterations of the test sets provided by FLORES200, using `fastText` skipgram models on a downstream language identification task (extrinsic evaluation). For evaluation, while F1 scores are popular in langID studies, they are hard to interpret and only have significant advantages when there is a class imbalance in the data distribution. We have selected a training and test set that is evenly distributed and is *not* imbalanced. Therefore, we opt for reporting top-1 accuracy since it is appropriate here and easier to interpret.

**Three Baseline Models** Our first baseline model (FLORES200) was trained on the raw language .dev files from FLORES200. We chose this as a baseline, given that it represents an easy and intuitive approach to training a language classification model, without any augmentation or modifications. Our second baseline (SEPARATE) keeps all script-language pairs separate during training and classification (Brown, 2012). That is, for 4 languages and 4 unique scripts, we'll end up with 16 total prediction classes. For reporting accuracies, we pool together results from all scripts for each language. Our third baseline (`WIKI`) is a language identification model pre-trained on Wikipedia, SETimes, and Tatoeba, boasting support for 176 languages (Joulin et al., 2017). Note that due to this large discrepancy in training data size, its performance will not be directly comparable to other models.

## 4 Results

We present our results for the Baseline, Flattening, Upscaling, and Noisy models here. In general, our script-agnostic models demonstrate good performance above the baselines on the transliterated test sets, and our methods comparable to traditional approaches on clean data.

### 4.1 Script Flattening

Under the Flattening experimental setup, even though certain languages have higher accuracies than others, each language appears to have comparable performance across scripts ( Table 4). For instance, Tamil sees ~80% accuracy on all flattened tests; in fact, each language's scores vary less than one percent when flattening to any given script. The uniformity across scripts suggests that any particular script does not play a major role in the models' decision-making. This matches and confirms our initial hypotheses, since there is no alternative script for the model to consider when evaluating language identity. Upon comparison with the baseline, our flattened models are far superior both in unconventional script scenarios, and when averaged across the four languages. In some cases, the baseline only classifies correctly 25% of the time, while our models consistently perform with over 90% average accuracy on the transliterated FLORES200 test set. With respect to individual language scores, the baseline classifies with slightly more accuracy when language and writing system match, but this is merely due to its heavy reliance on script, and does not speak to its overall performance. When script and language are not the same, the baseline is easily fooled; for example, in many cases it cannot classify even a single example correctly for certain languages.

**Interpretability Analysis** Interestingly, there is a difference in performance across the individual language scores for both models, where they correctly identify certain languages more often than others. For example, Malayalam scores near 100%, while Tamil is only correctly classified 80% of the time. To interpret differences in accuracy scores across languages, we utilize a game-theoretic metric, Shapley Additive Explanations, or SHAP (Lundberg and Lee, 2017), to compute global-level explanations across the training dataset. We focus on finding explanations for false positive features in Tamil sentences that have been predicted as Malayalam. We obtained translations for Tamil using Agarathi[4] and Google Translate, and for Malayalam using Google Translate and Olam[5]. Appendix Table 10 displays all the relevant words and characters in mispredicted Tamil sentences. While not all positively weighted words may have exact parallels in Malayalam, we think the score may come from positively correlated morphological features within the word itself, since Tamil and Malayalam share many word suffixes, prefixes, pluralization rules, prepositions etc. Our interpretability investigation revealed that this is due to presence of some positive MAL signal in TAM sentences, due to the lexical, semantic, and phylogenetic similarity of the two languages. This overlap causes a small number of sentences to be assigned a high probability of both TAM and MAL, with MAL winning by a slight margin. For more detailed results, plots, and explanations, please refer to Appendix §D

[4] `https://agarathi.com`. அகராதி/*agarathi* means dictionary in Tamil

[5] Malayalam Dictionary - `https://olam.in/`

| Scripts → <br> Languages ↓ | **Taml** <br> Baseline | <br> Flatten | **Knda** <br> Baseline | <br> Flatten | **Mlym** <br> Baseline | <br> Flatten | **Telu** <br> Baseline | <br> Flatten |
|---|---|---|---|---|---|---|---|---|
| TAMIL | 94.37 | 80.43 | - | 80.63 | - | 80.93 | - | 80.73 |
| KANNADA | - | 91.60 | 92.59 | 92.19 | - | 91.60 | - | 91.70 |
| MALAYALAM | 69.27 | 99.31 | 88.93 | 98.32 | 100.00 | 98.42 | 88.93 | 98.91 |
| TELUGU | - | 93.68 | - | 93.77 | - | 93.08 | 94.07 | 93.77 |
| AVERAGE | 40.91 | 91.25 | 45.28 | 91.23 | 25.00 | 91.01 | 45.75 | 91.28 |

Table 4: We find that no particular script is best suited to the flattening task and each script can allow for identification of the four Dravidian languages relatively faithfully. Although marginally, the Telugu script Flatten model performs best and so we include it in cross-domain experiments in 4.4. There is also a noticeable drop in performance for Tamil language, regardless of script (see Appendix D for interpretability analysis). Columns show scripts and rows indicate language. Baseline models are trained on all four languages in their original scripts and then tested on the transliterated flatten setups. We expect them to only predict the corresponding language for each script (and have -, meaning 0, for others), but we observe that they sometimes predict other languages too, despite not seeing them in the training corpus.

| | **WIKI** | | **FLORES200** <br> 3,988 | | **train - 25%** <br> 3,984 | | **train - 50%** <br> 7,968 | | **train - 75%** <br> 11,952 | | **train - 100%** <br> 15,952 | |
|---|---|---|---|---|---|---|---|---|---|---|---|---|
| | ORI | TRA | ORI | TRA | ORI | TRA | ORI | TRA | ORI | TRA | ORI | TRA |
| TAM | 100 | 25 | 94.37 | 23.59 | 48.02 | 48.84 | 77.96 | 78.04 | 91.8 | 92.02 | 95.26 | 95.16 |
| KAN | 100 | 25 | 92.59 | 23.15 | 74.41 | 74.18 | 89.62 | 90.02 | 92.69 | 92.76 | 95.06 | 95.06 |
| MAL | 100 | 25 | 86.78 | 95.85 | 95.41 | 99.11 | 97.83 | 99.7 | 99.68 | 99.7 | 99.65 | 99.65 |
| TEL | 100 | 25 | 94.07 | 23.52 | 47.23 | 46.89 | 92.49 | 92.86 | 94.37 | 94.47 | 95.36 | 95.41 |
| AVG | 100 | 25 | 95.26 | 39.26 | 66.38 | 66.33 | 89.80 | 89.69 | 94.64 | 94.73 | 96.35 | 96.32 |

Table 5: Transliteration of at least 75% of the data is required for Upscale models to perform at par with comparable baselines (FLORES200) on naturally occurring text. Additionally, these Upscale models also show high performance on transliterated test sets. The first two columns evaluate the `fastText` baselines on `WIKI` and FLORES200 datasets. The next four columns show Upscale models, trained on 25%, 50%, 75%, and 100% of the original training examples transliterated. The row underneath displays the amount of training data. Each model was tested on the original test set (ORI), without any transliterations, and a test set (TRA) with all examples transliterated to all scripts. Rows show language-specific langID performance.

### 4.2 Upscale

Our upscaled model performs quite well on the test sets, with over 96% accuracy (Table 5). Moreover, while it drastically outperformed the baseline on transliterated data, it scores higher on native script sentences as well. These results demonstrate that the model was able to correctly disentangle script and language using data augmentation.

**Comparison with Flattening** When comparing the Flattening results to Upscale, it is important to recognize that the latter model was trained on *four times* the amount of data, since we transliterated to all four scripts as opposed to flattening to a single script. Granted, the task was more complex as the model needs to handle 4 different writing systems per language. But, in order to nor-malize the effect of the number of examples, we also trained it using three variations of our training data: 25%, 50%, and 75% of the original examples transliterated. As expected, the 25% model performed much worse than the 100% model, and we saw improvements as we included more of the data. Interestingly, the results were only comparable to the Flattening model once we trained with at least 75% of the original examples. We suspect this is due to the difference between the number of cross-language examples and the number of cross-script examples. For instance, even though the 25% Upscaled model has nearly the same number of training examples as any of the Flattening models, many of these sentences are merely transliterated versions of each other, rather than full translations or original examples. This distribution appears to allow the model to become script-agnostic, but sacrifices the ability to identify languages in the process. This suggests that although Upscaling may perform better than Flattening overall, Flattening can perform similarly with *fewer* examples.

| | FLORES-TRA | FLORES-ORI | GLOT | UDHR | MCS350 | Avg |
|---|---|---|---|---|---|---|
| BASELINE (FLORES200) | 39.26 | 95.26 | 82.41 | 79.00 | 45.34 | 68.25 |
| 4-WAY PARALLEL | 96.32 | 96.35 | 81.67 | 77.54 | 44.79 | 79.33 |
| NON-PARALLEL | 94.39 | 94.37 | 84.61 | 83.86 | 51.76 | 81.80 |

Table 6: This table compares two Upscaled models, each trained on 997 examples per language, which are then transliterated to all scripts. One is trained on 4-way parallel data, and the other on examples that are not parallel from IndicCorp. The slight discrepancy of performance is likely a result of data from different domains. TRA for FLORES represents the test set that contains transliterations and ORI represents the default FLORES test set.

**Learning without $n$-way parallel data** It seems that Upscale models correctly ignore script in their decision-making process and so far, they have been trained on $n$-way parallel data; however, this could be a potential confounder. Therefore, we compare the performance of two script-upscaled models –one trained on 4-way parallel data, the other on non-parallel data– keeping the number of training examples per language constant for fairness. For non-parallel data, we use subsets of the monolingual corpora from IndicCorp for Telugu, Tamil, and Malayalam. We reuse the FLORES200 examples for Kannada, since these are not parallel to the data for the other three languages.

Our evaluation on the FLORES transliterated and clean test sets as well as all out-of-domain sets is in Table 6. The two models have largely similar results. The original 4-way parallel model does somewhat better on the FLORES test sets, and the non-parallel model has the better accuracy on average; however, these discrepancies can be expected due to the domain differences in data sources. Overall, it appears that both models are comparable and therefore using explicitly parallel data has a negligible effect.

### 4.3 Noisy Multi-Script

In the intra-sentence noise setup, performance varies to a large degree between the models, but accuracy distributions for each model stay relatively constant across test sets (Table 7). Our Script-Upscaled model is the best on average with over 99% accuracy, and the All-Noise model follows closely behind with a 98.82% score. Beyond these two, scores drop significantly to the 50-65% range, which is undesirable for a 4-class langID task.

This is likely explained by the size of the training sets. The Baseline, as well models with noise settings from 25 to 100, used data from four sets (one for each language) with varying script permutations. However, our All-Noise model was trained on a merged dataset consisting of sentences at *all* noise levels (i.e. four times the data). This is similar to the Script-Upscaled model that had access to each language's sentences transliterated to the four different scripts, and is likely what allowed the two models to perform so well. We believe that the Script-Upscaled model performed the best because it was consistently shown the same sentence in all four scripts, forcing it to become truly script-agnostic. The All-Noise model was able to do this to a large degree, but due to randomness and slight inconsistencies in permutations, it likely was not able to completely disregard script in its decision-making process. Therefore, script-mixing *within* sentences seems to be an extremely challenging setup for models and requires data augmentation for reasonable performance.

| Data | Language | Baseline | N@25 | N@50 | N@75 | N@100 | N@all | Upscale |
|---|---|---|---|---|---|---|---|---|
| CLEAN | Tamil | 23.59 | 40.19 | 14.95 | 42.81 | 26.75 | 93.08 | 95.26 |
| | Kannada | 23.15 | 76.38 | 58.75 | 77.32 | 67.27 | 93.33 | 95.16 |
| | Malayalam | 86.78 | 94.54 | 99.93 | 95.11 | 99.51 | 99.63 | 99.70 |
| | Telugu | 23.52 | 51.63 | 40.07 | 44.64 | 51.14 | 94.89 | 95.45 |
| all | Tamil | 40.77 | 36.86 | 14.82 | 39.66 | 25.55 | 99.77 | 100.00 |
| | Kannada | 39.72 | 77.02 | 56.24 | 78.59 | 65.25 | 99.02 | 99.14 |
| | Malayalam | 86.94 | 96.34 | 99.97 | 96.24 | 99.57 | 99.90 | 99.95 |
| | Telugu | 42.40 | 52.70 | 38.71 | 43.32 | 52.47 | 99.47 | 99.77 |
| AVG | * | 50.27 | 65.72 | 52.60 | 64.52 | 60.79 | 98.82 | 99.16 |

Table 7: Even after introducing transliteration noise at different levels within sentences, the N@all and Upscale models are competitive implying that we can use word-level script-mixing without sacrificing performance. The table has been abridged due to space constraints, but an extended version with results for 25, 50, 75, and 100% noise-levels is in Appendix Table 9. N@25,50,75,100 and the baseline models were trained with 3988 sentences per class. The Upscale and N@all models (last two columns) were trained with 15952 sentences per class and are therefore more comparable with each other. The baseline was trained on original FLORES200 data.

| Test Dataset → | **FLORES200** | **GLOT** | **UDHR** | **MCS350** | **AVERAGE** |
|---|---|---|---|---|---|
| Test Set Size → | 4048 | 3934 | 285 | 15000 | 5817 |
| BASELINE (WIKI) | 100.00 | 99.96 | 100.00 | 71.75 | 92.93 |
| BASELINE (SEPARATE) | 25.00 | 24.92 | 20.35 | 25.00 | 23.81 |
| BASELINE (FLORES200) | 95.26 | 82.41 | 79.00 | 45.34 | 75.50 |
| FLATTEN (TELU) | 91.28 | 43.18 | 44.56 | 33.95 | 53.24 |
| UPSCALE (16K) | 96.35 | 81.67 | 77.54 | 44.79 | 75.09 |
| NOISE (ALL) | 95.41 | 80.19 | 76.14 | 43.41 | 73.79 |

Table 8: We share three `fastText`-based baseline models (trained on FLORES200, separate language and script classes, and Wikipedia) along with the best model from each of our 3 experimental setups (upscale, flatten, noise). We test them on out of domain data to test domain transfer of the learned embeddings. Overall, the UPSCALE (16K) and NOISE (ALL) models have comparable performance to BASELINE (FLORES200) demonstrating that the multi-script training doesn't lead to a significant degradation in performance on the languages' naturally occurring native scripts. Note that the `WIKI` model is trained on all of Wikipedia, and therefore its performance is not directly comparable to any of the other models. The SEPARATE baseline performs the poorest, likely due to the low amount of data required for a 16-way classification task.

### 4.4 Cross-Domain Performance

A comparison of our models on the clean FLORES200 test set, as well as out-of-domain sets is in Table 8. The FLORES200 BASELINE performs well in-distribution and on similar long-length GLOT and UDHR datasets, but poorly on MCS350 (children's stories domain and shorter sentences). The `WIKI` baseline is better than the FLORES200 baseline across all datasets, showing that is has built a better representation space for the languages. The UPSCALE (16K) and NOISE (ALL) models have comparable performance to BASELINE (FLORES200), demonstrating that the multi-script training does not lead to a significant degradation in performance on the languages' conventional/native scripts. The FLATTEN algorithm naturally performs poorly compared to the other models in this setting since it is only exposed to one script. Therefore, it may not be a practical choice for real-world language identification.

## 5 Discussion

The results demonstrate that all of our script-agnostic language identification models (Flattening, Noise, and Scipt-Upscaled) perform well above the baselines on examples that utilize a non-standard script. In certain cases where data is in native script, our baseline models can surpass some script-agnostic ones; this is likely because the baselines use script as a basis for determining language ID. The All-noise model showed very good performance, and we suspect it remains second to the Upscaled setting primarily due to the variability of the training data. Unlike the Upscaled model, it may not see every example transliterated to all scripts, and thus may not become completely agnostic of script. However, it is a strong contender and its performance on other downstream tasks and the quality of its learned representations should be evaluated in future work when scaling to a larger number of languages and scripts.

In the practical setting, our models –especially Script-Upscaled– appear to be a reasonable alternative to current language identification systems. Additionally, it is likely that had we trained an Upscaled model on Wikipedia, we would have seen results that matched the `WIKI` baseline on noiseless data. The large amount of storage and computational power for this endeavor, in addition to potential challenges in transliterating to so many scripts, would have been beyond the scope of our current work. However, future work should carefully explore creation of script-agnostic `WIKI` langID models as well. Our Upscaling approach is relatively straightforward, and requires no more examples than for a standard language identification system. Since transliteration can be done automatically and cheaply, our final proposal is a script-based data-augmentation process for complete sentences and within sentences. When expanding to other languages and scripts, lossy transliteration quality in non-Indic systems may be a challenge, and we recommend using the International Phonetic Alphabet (IPA) as a bridge for high-quality and natural transliteration.

## 6 Related Work

Previous work has demonstrated that script barriers discourage transfer learning from high-resource languages into low-resource languages' representation spaces, especially for Neural Machine Translation (Muller et al., 2021; Anastasopoulos and Neubig, 2019). Moreover, script diversity negatively impacts low-resource languages disproportionately because their training data is often of poor quality and smaller in size (Pfeiffer et al., 2021). Consequently, researchers have focused on transliteration, romanization, phonetic representation etc. to reduce vocabulary sizes and allow lexical sharing between languages with different writing systems (Amrhein and Sennrich, 2020).

Another common approach relies on existing pre-trained models and fine-tuning them with different transliterated versions of the originally supported languages (Muller et al., 2021; Dhamecha et al., 2021). This is an instance of the common hierarchical pipeline (Goutte et al., 2014; Lui et al., 2014; Bestgen, 2017) or fine-tuning-based approach for language identification (Jauhiainen et al., 2018; Agarwal et al., 2023; Ahmadi et al., 2023a). Most recently, Moosa et al. (2023) conducted a study on effects of transliteration on multilingual language modeling, which focused on two kinds of models: a multi-script model with native scripts of each language (matching our BASELINE setup) and a uni-script model with only one script for all languages (similar to our FLATTEN setup).

As a natural extension of their work, we also consider UPSCALE and NOISE setups for Dravidian languages, as described in §3. Unlike their work, we do not fine-tune on downstream tasks, but instead focus on including the transliteration in the original training data to give the model the ability to handle non-native scripts without losing performance on the original script. Moreover, our work is not only motivated from a lexical-sharing and transfer-learning perspective, but is grounded with the aim of supporting synchronic and diachronic digraphia adequately in NLP applications and tasks.

## 7 Conclusion

We introduce and evaluate three new kinds of language identification models that are script-agnostic. All of our systems have been shown to outperform the baseline on examples that are not written in the standard script. Two of our models (Upscaled and All-Noise) perform especially well on both clean and transliterated data. Our methods may provide a reasonable alternative to training language identifiers that can correctly classify text based on the language used, rather than the script in which it is written. Future work should expand to include more languages and scripts, as well as performing thorough intrinsic evaluation on the learned embeddings to determine if these would be effective on other downstream tasks.

## Limitations

**Extending to a larger set of languages** We note that our models were only trained and evaluated using the four major Dravidian languages - Tamil, Telugu, Malayalam, and Kannada. Extending the successful experiments (upscale and all-noise) to a larger number of writing systems may prove challenging in terms of computational resources and dataset sizes. Data loss associated with script conversion and non-phonetic scripts is a likely challenge (and *potential* limitation) when we scale our approach to more scripts.

**Unknown Scripts** Note that our approach helps bring script-agnosticism to scripts included during training time. The model will still struggle with unknown writing systems, and for this, we will need to scale to an extremely large number of writing systems, which we leave for future work.

**Data loss due to script-conversion** Most Indic scripts have a 1:1 phonetic mapping between graphemes, but there may still be letters that are not mapped accurately (truly unique sounds in certain languages). In our study, three of the four scripts have direct phonetic mappings, while only one (Tamil) includes aspirated sounds that are not translatable to the other writing systems. This means that two different scripts representing the same word can have two different character distributions.

## Ethics Statement

Languages may be written in non-native scripts to obfuscate their presence on the internet, and the use script-agnostic embeddings would be able to discover and accurately identify such text during web crawls. This may have some downstream privacy and surveillance related concerns that are out of scope for this work. Currently, our pilot study uses the FLORES200 dataset to train embeddings, but in the future, a larger corpora such as Wikipedia, CommonCrawl, or other publicly crawled data can be used, which may bring with it several concerns around data ownership and copyright.

## Acknowledgements

This work was generously supported by the National Endowment for the Humanities under award PR-276810-21, by the National Science Foundation under award IIS-2327143, and by a Sponsored Research Award from Meta. Joshua Otten is also supported by the Presidential Scholarship awarded by the George Mason University Graduate Division. Computational resources for experiments were provided by the Office of Research Computing at George Mason University (URL: `https://orc.gmu.edu`).

## A Out-of-Domain Datasets

1. **FLORES200**: Open-source $n$-way parallel dataset consisting of sentences from 842 web articles, translated into a large number of languages (Costa-jussà et al., 2024; Goyal et al., 2022; Guzmán et al., 2019). Each language's example are in the same order, and are separated into `.dev` and `.devtest` files, containing 997 and 1012 sentences, respectively.

2. **GlotStoryBooks**[6]: Open-licensed curated library of books (Kargaran et al., 2023) from a variety of sources in 176 languages (Yankovskaya et al., 2023; Ogundepo et al., 2023). Each sample contains a sentence along with its language identifier and script.

3. **UDHR (Universal Declaration of Human Rights)**: We use Kargaran et al. (2023)'s public domain preprocessed version of the UDHR dataset, where each sample is a paragraph along with a language identifier. The authors removed errors and formatting issues in the original UDHR data and made this clean version available[7].

4. **MCS-350**: Multilingual Children's Stories dataset, released by Agarwal et al. (2023), contains over 50K children's stories curated primarily from two sources - African Storybooks Initiative and Pratham Storyweaver, both open-source story repositories for African and Indian languages respectively. For our experiments, we use the monolingual data files available on the authors' GitHub repository[8] for Tamil, Malayalam, Kannada, and Telugu. Compared to UDHR, the sentences are relatively smaller in length since they are not from the legal domain, and unlike GlotStoryBooks, the authors don't apply any length-based filtering to the curated stories.

5. **IndicCorp**[9]: Monolingual, sentence-level corpora for English and 11 Indian languages from the Dravidian and Indo-Aryan families (Kakwani et al., 2020). It consists of 8.8 billion tokens and is sourced mostly from Indian news crawls (articles, blog posts, magazines), though it also takes data from the OSCAR corpus.

[6] https://huggingface.co/datasets/cis-lmu/GlotStoryBook
[7] https://huggingface.co/datasets/cis-lmu/udhr-lid
[8] https://github.com/magarw/limit
[9] https://paperswithcode.com/dataset/indiccorp

## B Brief Language Profiles

1. Tamil (`tam`), a Southern-Dravidian language, is spoken by over 80 million people and is an official language in Sri Lanka, the Indian states of Tamil Nadu and Puducherry, and of the Indian Constitution's Eighth Schedule. It is curently most widely written in the Tamil abugida - தமிழ் எழுத்து (*tamizh ezhuttu*).

2. Telugu (`tel`), a South-Central Dravidian language, is spoken by about 100 million people and is the most spoken Dravidian language. It is also an Eighth Schedule language of the Indian Constitution and is official in the Indian states of Andhra Pradesh, Telangana, and Puducherry (Yanam). It is written in Telugu abugida - తెలుగు లిపి (*telugu lipi*)

3. Malayalam, (`mal`), another Southern-Dravidan language is the smallest language from our selection, spoken by about 40 million people in Southern India. It is an Eighth Schedule language and is official in the southernmost Indian state of Kerala. It is written in the Malayalam abugida - മലയാളം അക്ഷരങ്ങൾ (*malayalam aksharangal*).

4. Kannada (`kan`), also a member of the Southern-Dravidian language subfamily, is spoken by about 60 million people, mostly within India. It is an official language of the Indian Constitution's eighth schedule and is the sole official language of Karnataka state. It is widely written in Kannada script, which is closely related to the Telugu script and is also an abugida, but diverged around 1300 CE - ಕನ್ನಡ ಅಕ್ಷರಮಾಲೆ (*kannada aksharamale*).

## C Noise-Experiments Extended Results

## D Interpreting Flattening Results

The default baseline (in-distribution) is a `fastText` model trained on FLORES200 data, keeping the languages in their original scripts without any transliterations. For the *flattening* experiments, we project all data to one script at a time. Since the test data is flattened to a single

| Data | Language | Baseline | N@25 | N@50 | N@75 | N@100 | N@all | Upscale |
|---|---|---|---|---|---|---|---|---|
| CLEAN | Tamil | 23.59 | 40.19 | 14.95 | 42.81 | 26.75 | 93.08 | 95.26 |
| | Kannada | 23.15 | 76.38 | 58.75 | 77.32 | 67.27 | 93.33 | 95.16 |
| | Malayalam | 86.78 | 94.54 | 99.93 | 95.11 | 99.51 | 99.63 | 99.70 |
| | Telugu | 23.52 | 51.63 | 40.07 | 44.64 | 51.14 | 94.89 | 95.45 |
| 25 | Tamil | 35.05 | 36.25 | 14.20 | 39.98 | 25.08 | 99.90 | 100.00 |
| | Kannada | 31.38 | 77.21 | 56.93 | 78.82 | 64.76 | 99.40 | 99.60 |
| | Malayalam | 85.74 | 95.58 | 99.90 | 96.18 | 99.80 | 99.90 | 100.00 |
| | Telugu | 36.18 | 52.66 | 38.29 | 43.82 | 51.96 | 99.10 | 99.90 |
| 50 | Tamil | 38.87 | 36.86 | 14.30 | 39.68 | 26.38 | 99.70 | 100.00 |
| | Kannada | 41.84 | 77.30 | 55.83 | 79.23 | 66.67 | 99.29 | 99.59 |
| | Malayalam | 86.00 | 96.48 | 100.00 | 96.17 | 96.17 | 99.90 | 99.90 |
| | Telugu | 43.50 | 52.47 | 38.67 | 42.50 | 52.77 | 99.40 | 99.70 |
| 75 | Tamil | 45.01 | 37.34 | 14.83 | 39.35 | 25.93 | 99.80 | 100.00 |
| | Kannada | 44.08 | 76.34 | 56.22 | 78.77 | 64.91 | 98.79 | 98.89 |
| | Malayalam | 88.56 | 96.86 | 100.00 | 95.85 | 99.39 | 100.00 | 100.00 |
| | Telugu | 47.61 | 52.59 | 39.19 | 43.65 | 52.79 | 99.59 | 99.70 |
| 100 | Tamil | 44.21 | 36.99 | 15.96 | 39.63 | 24.80 | 99.70 | 100.00 |
| | Kannada | 41.19 | 77.23 | 55.97 | 77.53 | 64.68 | 98.58 | 98.48 |
| | Malayalam | 87.46 | 96.43 | 100.00 | 96.74 | 99.39 | 99.80 | 99.90 |
| | Telugu | 42.35 | 53.09 | 38.70 | 42.96 | 52.38 | 99.80 | 99.80 |
| all | Tamil | 40.77 | 36.86 | 14.82 | 39.66 | 25.55 | 99.77 | 100.00 |
| | Kannada | 39.72 | 77.02 | 56.24 | 78.59 | 65.25 | 99.02 | 99.14 |
| | Malayalam | 86.94 | 96.34 | 99.97 | 96.24 | 99.57 | 99.90 | 99.95 |
| | Telugu | 42.40 | 52.70 | 38.71 | 43.32 | 52.47 | 99.47 | 99.77 |
| AVG | * | 50.27 | 65.72 | 52.60 | 64.52 | 60.79 | 98.82 | 99.16 |

Table 9: Even after introducing noise at all levels, the N@all and Upscale models are competitive implying that we can both use the word-level script-mixing without sacrificing performance on clean or noisy data. Among the noise@25,50,75 settings, we observe that 50% and 100% noise have drastic impact on classification accuracy for $\geq$ 2 languages. N@25,50,75,100 and the baseline models were trained with 3988 sentences per class. The Upscale and N@all models were trained with 15952 sentences per class and are therefore more comparable with each other. The baselinen was trained on FLORES200 data.

**Algorithm 1** Synthetic Noise Within Sentences

**for** $noise = 25, 50, 75, 100$ **do**
    **for** $lang = tam, kan, mal, tel$ **do**
        **for** *sent* $= 0, 1, \ldots..N$ **do**
            Choose 1 base script
            Choose noise% words to transform
            **for** *index in chosen indices* **do**
                *nonbase* = Chose new script
                Transform word into *nonbase*
    Save transformed data at *noise*-level
Merge-save sentences at all *noise* levels into a new file for the *all*-noise setting script, we would expect the model to only predict the language that is representative of the writing system. For instance, the baseline model would predict Tamil when it's shown data from any language in the Tamil script. But, we find that the models (trained on data in 4 different scripts and languages) tend to default to a Malayalam prediction for sentences that it knows are not Tamil (Table 4). This can be seen by the presence of a Malayalam signal across experiments for all 4 projection scripts. It also seems that several Malayalam sentences are being misclassified as Tamil (as evident by the less-than-100% accuracy for the Malayalam row for non-Malayalam

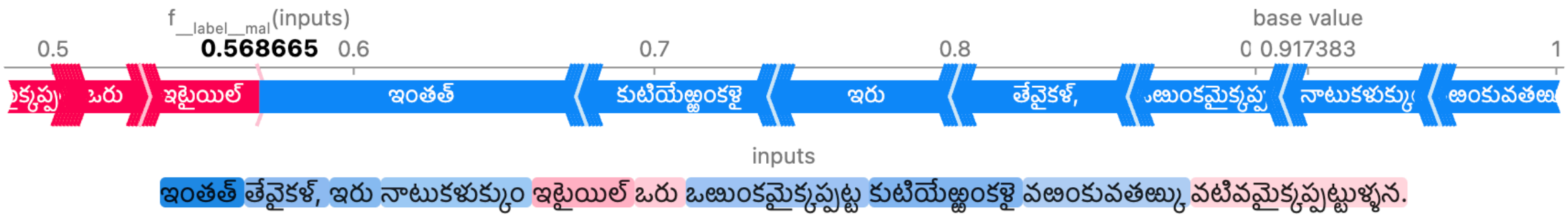

Figure 2: Example: Sentence 0's SHAP visualization for gold TAM sentence and weights when predicted class is MAL. Red indicates positive signal for MAL (unwanted) and blue indicates negative signal for MAL (wanted).

scripts).

For the Upscale experiments (Table 5), we find that the Wikipedia pre-trained model does not have the same bias towards Malayalam as our model, and instead is perfectly fit to each language's writing system (100% and 25% accuracy on Original and Transliterated data). The custom-trained FLORES200 baseline, on the other hand, has similar performance (between 86-94% for Original and 23% for Transliterated). We observe the Malayalam-defaulting phenomenon here as well, and it is likely that the model is over-predicting Malayalam, treating it as an "other" prediction bucket.

For Noise experiments (Table 7), we observe similar performance by the FLORES200 baseline as on the Upscaling experiments. However, the accuracy for non-Malayalam languages seems to increase as we increase the amount of noise. To interpret differences in accuracy scores across languages, we utilize a game-theoretic metric, Shapley Additive Explanations, or SHAP (Lundberg and Lee, 2017), to compute global-level explanations across the training dataset for all 4 languages. As discovered in 4.1, we find that Tamil receives a significantly lower accuracy (around 80%) compared to the other 3 languages, especially compared to Malayalam (95%+). Therefore, we focus on finding explanations for false positive features in Tamil sentences that have been predicted as Malayalam. Readers should note that Tamil and Malayalam are closely related since they were the most recent to diverge from each other among the four major Dravidian languages (around the 9th century CE). Therefore, there are substantial vocabulary and grammatical similarities between them.

Table 10 displays all the relevant words and characters in mispredicted Tamil sentences. We obtained translations for TAM using Agarathi[10] and Google Translate, and for MAL using Google Translate and Olam[11]. While not all positively weighted words may have exact parallels in Malayalam, we think the score may come from positively correlated morphological features within the word itself, since Tamil and Malayalam share many word suffixes, prefixes, pluralization rules, prepositions etc. It is worth noting that our interpretability study revealed that for the flattened script condition, the `fastText` trained models always predict MAL as default. This is not inherently bad because we still receive over 90% accuracy for MAL, KAN and TEL, indicating that the models find sufficient non-MAL signal in the sentence when it's present. However, for TAM, we saw that there was a 10% gap in performance (i.e TAM prediction accuracy stayed around 80%). Our interpretability investigation revealed that this is due to presence of some positive MAL signal in TAM sentences, due to the lexical, semantic, and phylogenetic similarity of the two languages. This overlap causes a small number of sentences to be assigned a high probability of both TAM and MAL, with MAL having the maximum since it is the default prediction being downscored.

Results and all graphs from the Interpretability Jupyter notebook have been attached below. It shows the sentence-level explanations for each of the Tamil sentences that were misclassified in the training set with a small margin.

[10] https://agarathi.com. அகராதி/*agarathi* means dictionary in Tamil

[11] Malayalam Dictionary - https://olam.in/

| Sent | TAM in TELU script | Weight | Transliteration | MAL | TAM |
|---|---|---|---|---|---|
| WORDS | | | | | |
| 0 | **ఇటైయిల్** | 0.039 | *itaiyil* | during | in between |
| 0 | **ఒరు** | 0.025 | *oru* | a, an | a, an |
| 0 | వటివమైక్కప్పట్టుళ్ళన | 0.021 | *vativamaikkuppattullana* | shaped | are designed |
| 1 | వఴాంకప్పట్టతు | 0.041 | *vazhaankappattathu* | indulgence | provided |
| 2 | **ఇల్లై** | 0.021 | *illai* | no, not | no, not, ain't |
| 3 | **చిఴియవై!** | 0.031 | *chizhiyavai* | small ones | small ones |
| 4 | నిఴువప్పట్టతు | 0.039 | *nizhuuvappattathu* | | established |
| 5 | వరుక్కైక్కు | 0.059 | *varukkaikku* | | to visit |
| 5 | **ఒరు** | 0.058 | *oru* | a, an | a, an |
| 5 | **వఴంకాతు.** | 0.029 | *vazhankaathu* | don't give in | doesn't provide |
| 6 | పతివాకిన. | 0.033 | *pativaakina* | regularly | were recorded. |
| 7 | **ఒరు** | 0.035 | *oru* | a, an | a, an |
| 7 | చమైక్కప్పటుకిఴతు. | 0.024 | *chamaikkappatukizhathu* | | is being cooked |
| 8 | ఆతరవళిక్కవిల్లై. | 0.043 | *aatharavalikkavillai* | | not supported |
| CHARACTERS | | | | | |
| 0 | వటివమైక్కప్పట్టు**ళ్ళన** | 0.052 | *vativamaikkappattu**llana*** | | |
| 0 | **ఒరు_** | 0.037 | ***oru_*** | a, an | one |
| 0 | కు**టి**యేఴ్ఴంకళై | 0.037 | *ku**ti**yezzhankalai* | above | |
| 0 | ఇ**టైయిల్** | 0.035 | *it**aiyil*** | in | |
| 1 | వఴంకప్ప**ట్టత** | 0.102 | *vazhankappa**ttatha*** | suffix | suffix |
| 1 | కు**టి**నీర్ | 0.033 | *ku**ti**niir* | above | |
| 1 | అవర్క**ళు**క్కు | 0.024 | *avarku**lu**kku* | to them | they |
| 2 | కు**టి**యిరుప్పినుళ్ | 0.152 | *ku**ti**yiruppinul* | above | |
| 3 | చి**ఴియవ** | 0.125 | *chiz**hiyava*** | small ones | small ones |
| 4 | ఉరు**వా**క్కుం | 0.033 | *uru**vaa**kkum* | emerge | create |
| 5 | ని**ఴు**వప్పట్ట**త** | 0.112 | *niz**hu**vappatta**tha*** | | |
| 6 | **క్కు_** | 0.079 | ***kku_*** | | |
| 6 | వఴంకా**త** | 0.045 | *vazhankaa**tha*** | | |
| 6 | **_ఒరు** | 0.037 | ***_oru*** | a, an | one |
| 7 | పతి**వాకిన** | 0.119 | *pati**vaakina*** | | |
| 7 | మలై**యి**న్ | 0.022 | *malai**yi**n* | | |
| 8 | చమైక్కప్పటుకి**ఴత** | 0.048 | *chamaikkappatuki**zhatha*** | | |
| 8 | **కుఴి** | 0.035 | ***kuzhi*** | pit | pit |
| 9 | చే**ర్ప్ప**తై (ర్ ) | 0.035 | *chee**rppa**thai (r)* | | |

Table 10: Words and characters that have a positive Malayalam explanation weight of $> 0.02$ for ground-truth Tamil sentences. All sentences under consideration had a difference of $> 0.15$ between the Tamil and Malayalam classes. We pick this threshold since it gives us Tamil sentences that have a high-enough Malayalam signal (or low Tamil signal) causing the classifier to mispredict.

## Character Level Explanations > 0.15

In [ ]:
```
ix_array
```

Out[ ]:
```
array([ 906, 1113, 1395, 1687, 2080, 2108, 2224, 2270, 2801])
```

In [ ]:
```
for i in ix_array:
    shap.plots.text(shap_values[i])
```

outputs

__label__tam __label__kan __label__mal __label__tel

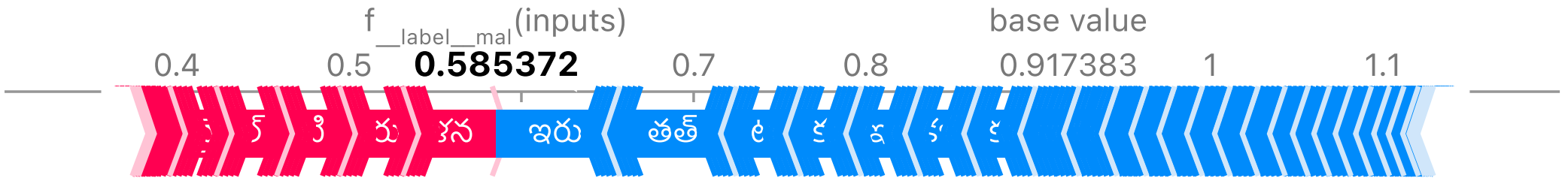

inputs

ఇంతత్ తేవైకళ్, ఇరు నాటుకళుక్కుం ఇట్టైయిల్ ఒరు ఒటుంకమైక్కప్పట్ట కుటియేట్టంకళై వటంకువతఱ్కు
వటివమైక్కప్పట్టుళ్ళన

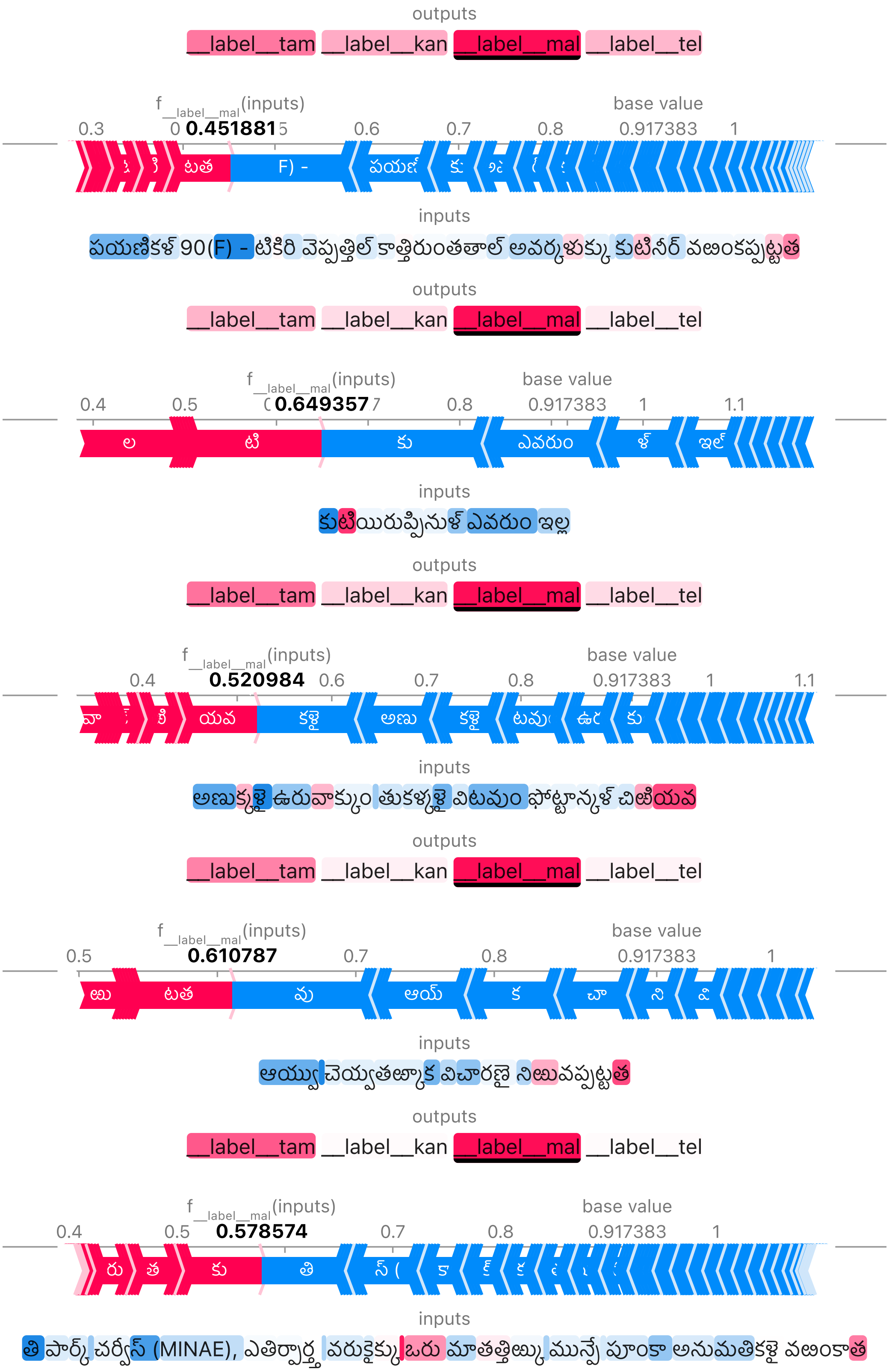

outputs
__label__tam __label__kan __label__mal __label__tel
f__label__mal(inputs)
0.451881
base value
0.917383
0.3 0.6 0.7 0.8 1
inputs
పయణికళ్ 90(F) - టికిర వెప్పత్తిల్ కాత్తిరుంతతాల్ అవర్కళుక్కు కుటినీర్ వఴంకప్పట్టత
outputs
__label__tam __label__kan __label__mal __label__tel
f__label__mal(inputs)
0.649357
base value
0.917383
0.4 0.5 0.8 1 1.1
inputs
కుటియిరుప్పినుళ్ ఎవరుం ఇల్ల
outputs
__label__tam __label__kan __label__mal __label__tel
f__label__mal(inputs)
0.520984
base value
0.917383
0.4 0.6 0.7 0.8 1 1.1
inputs
outputs
__label__tam __label__kan __label__mal __label__tel
f__label__mal(inputs)
0.610787
base value
0.917383
0.5 0.7 0.8 1
inputs
outputs
__label__tam __label__kan __label__mal __label__tel
f__label__mal(inputs)
0.578574
base value
0.917383
0.4 0.5 0.7 0.8 1
inputs

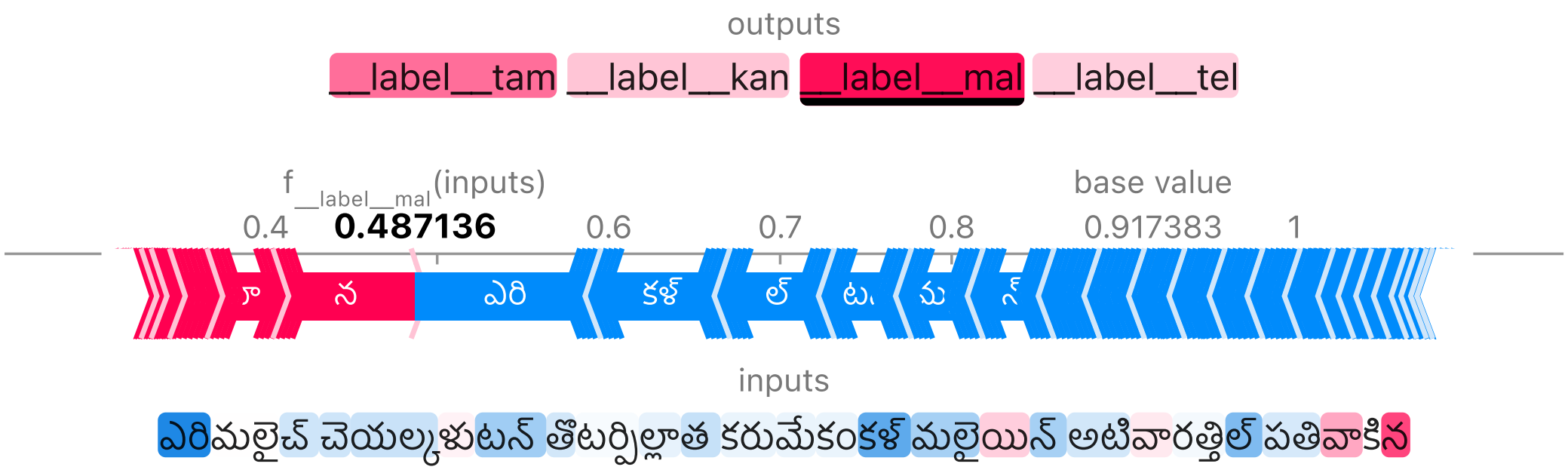

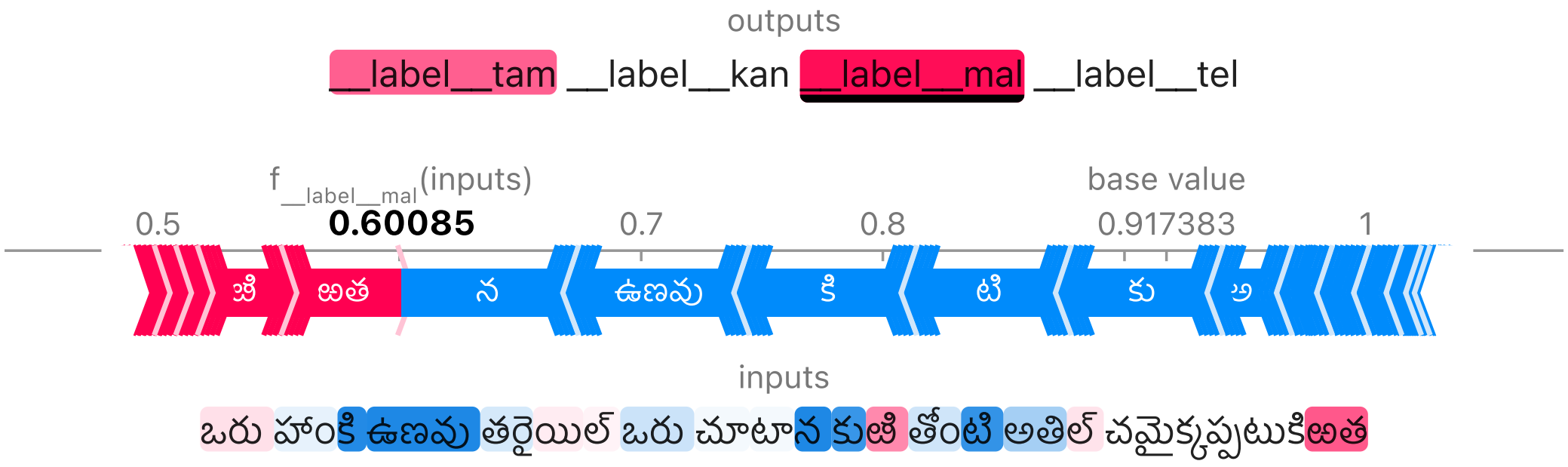

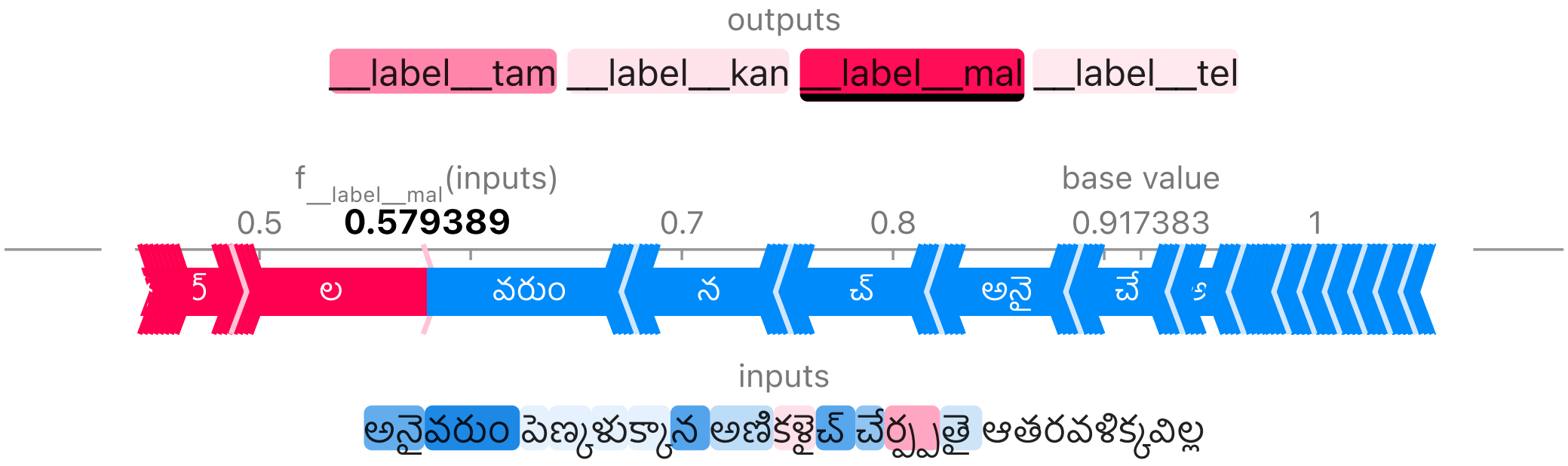

In [ ]:

```
for i in ix_array:
    shap.plots.bar(shap_values[i][:,2], max_display=20)
```

## Explanations > 0.15

In [ ]:
```python
for i in ix_array:
    shap.plots.text(shap_values[i])
```

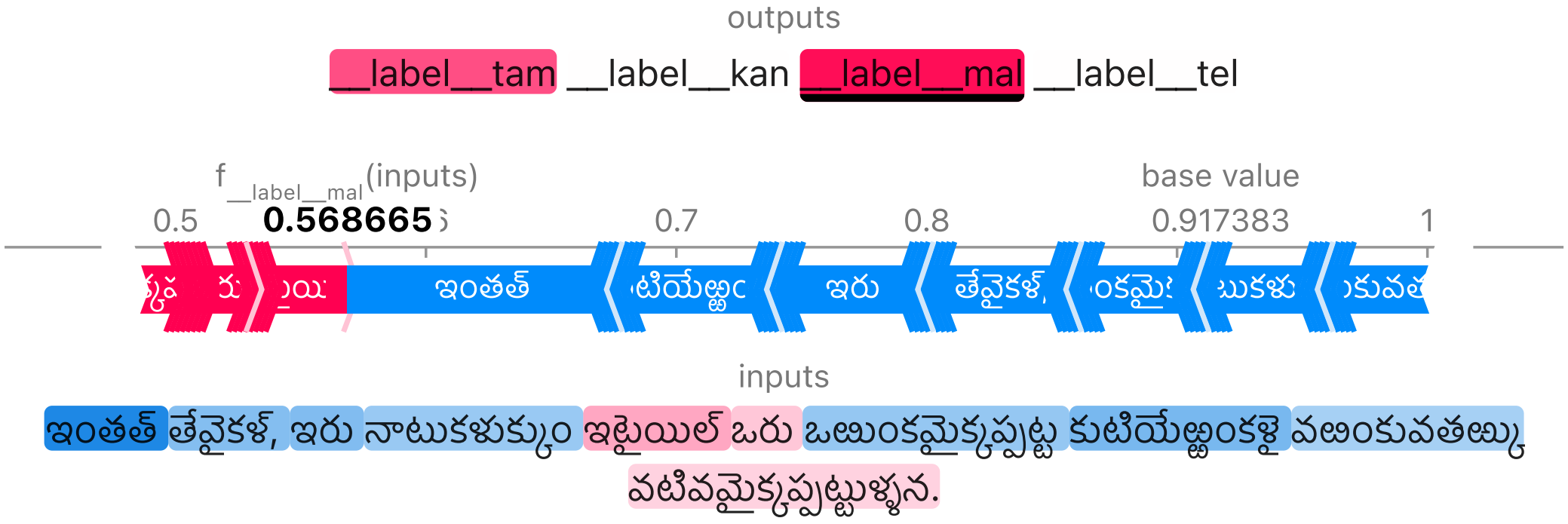

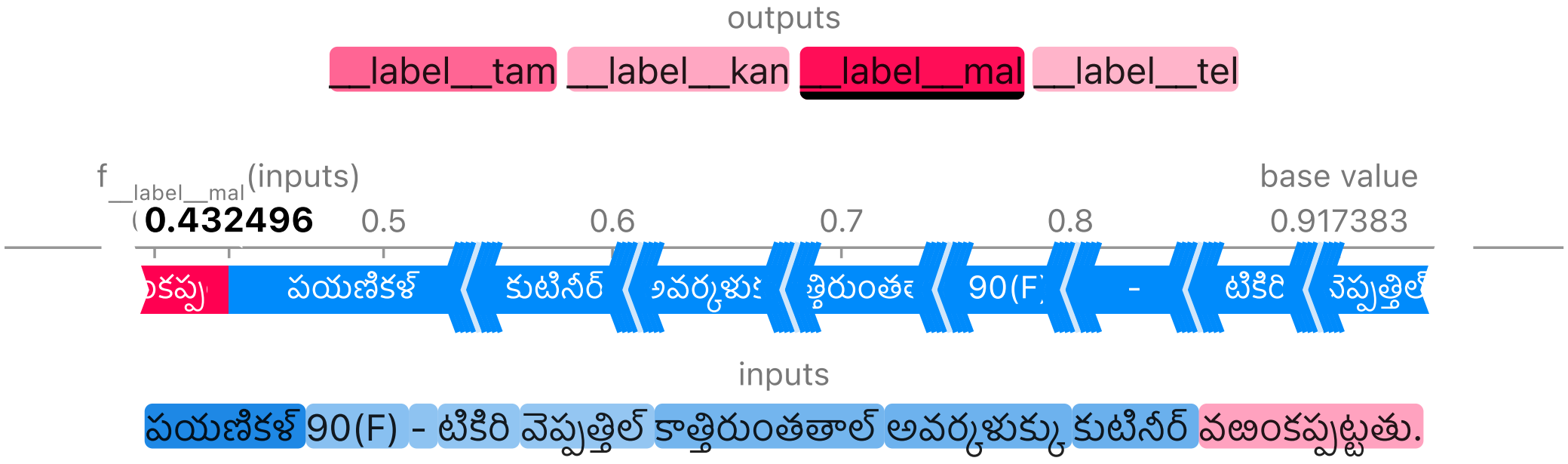

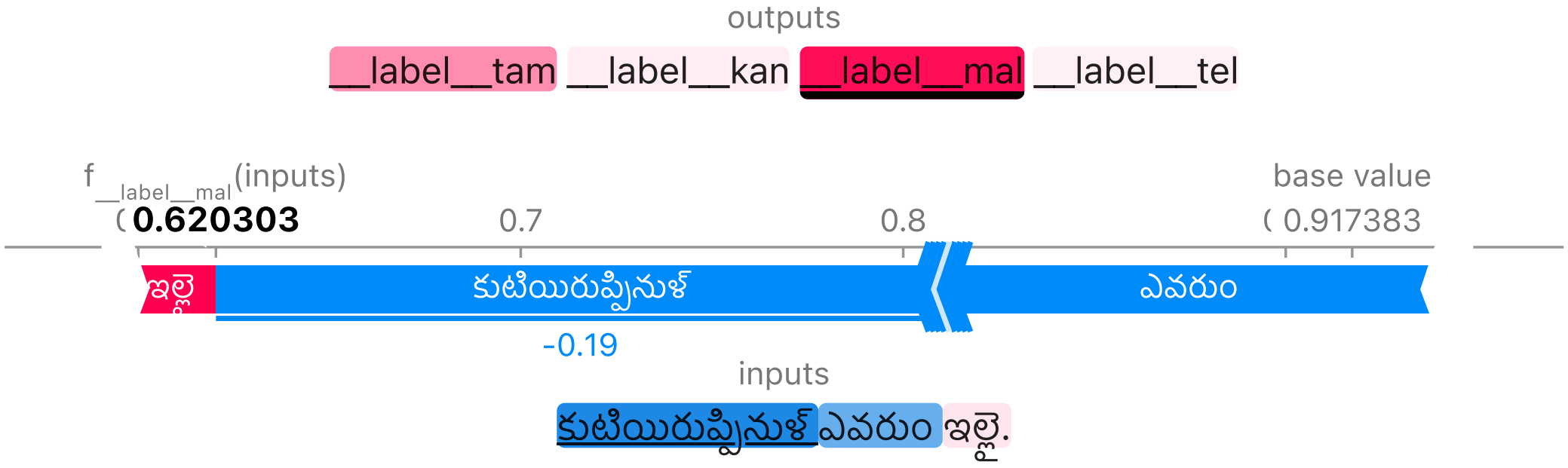

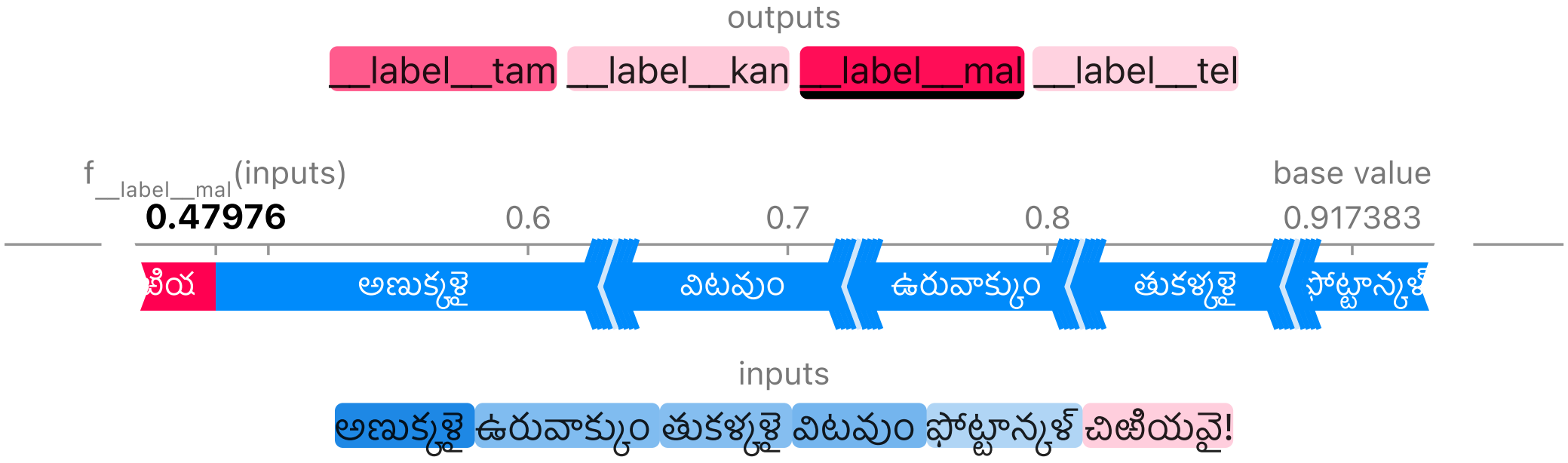

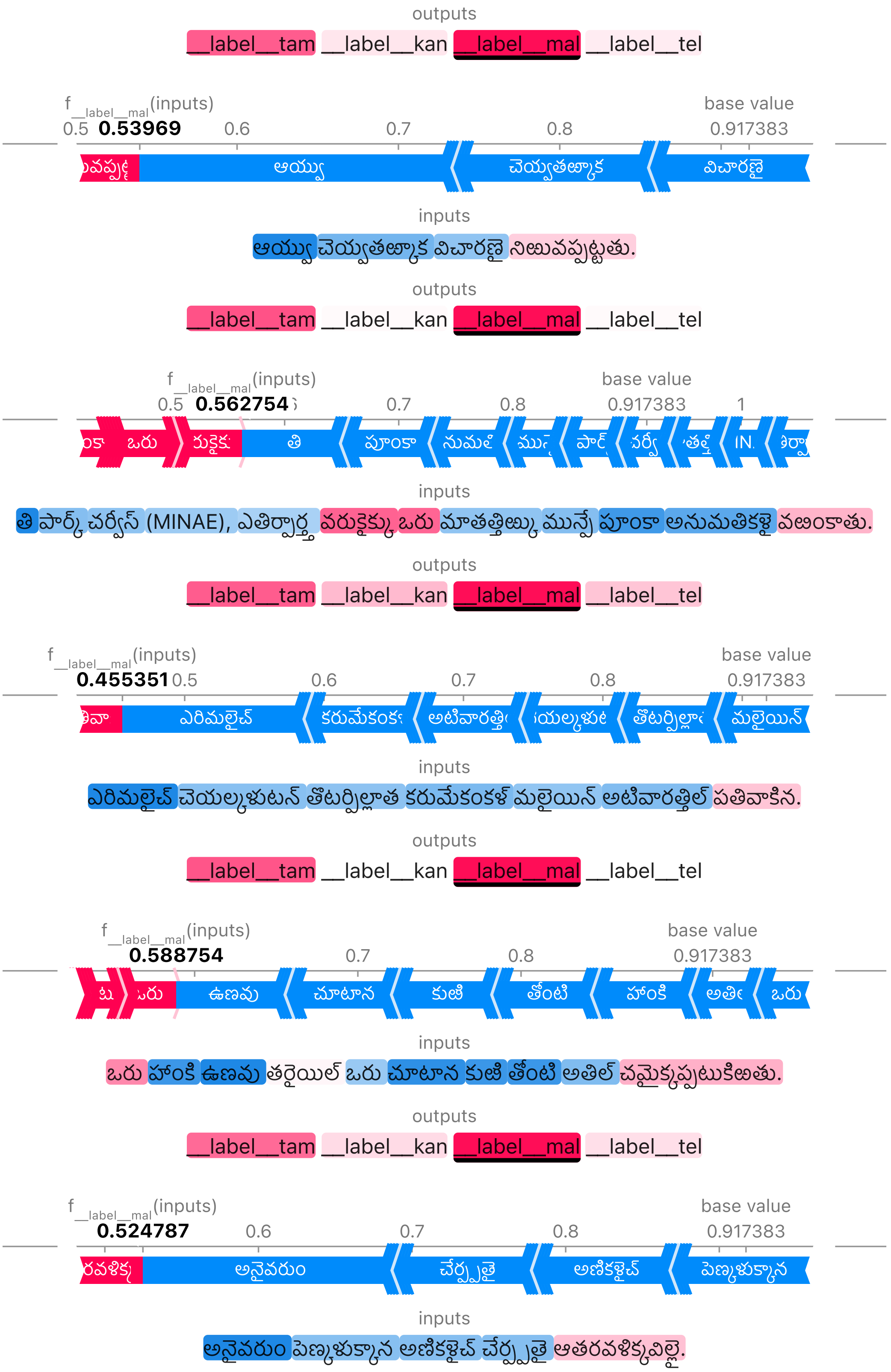